\documentclass[fleqn,10pt]{wlscirep}

\usepackage[utf8]{inputenc}
\usepackage[T1]{fontenc}
\usepackage{xcolor}
\usepackage{soul}
\usepackage{hyperref}
\usepackage{pdfpages}
\title{\color{black}
Artificial Intelligence-Based Methods for Fusion of Electronic Health Records and Imaging Data\thanks{This is pre-print of paper accepted for publication in Scientific Reports. Cite the final version from Nature Scientific Reports.\\ Email: haali2@hbku.edu.qa, zshah@hbku.edu.qa} }

\author[1]{Farida Mohsen}
\author[1]{Hazrat Ali}
\author[1,2]{Nady El Hajj}
\author[1,*]{Zubair Shah}
\affil[1]{College of Science and Engineering, Hamad Bin Khalifa University, Qatar Foundation, 34110 Doha, Qatar}

\affil[2]{College of Health and Life Sciences, Hamad Bin Khalifa University, Qatar Foundation, 34110 Doha, Qatar}

\affil[*]{Correspondence to: Zubair Shah}

\begin{abstract}
Healthcare data are inherently multimodal, including electronic health records (EHR), medical images, and multi-omics data. Combining these multimodal data sources contributes to a better understanding of human health and provides optimal personalized healthcare. The most important question when using multimodal data is how to fuse them - a field of growing interest among researchers.  Advances in artificial intelligence (AI) technologies, particularly machine learning (ML), enable the fusion of these different data modalities to provide multimodal insights. To this end, in this scoping review, we focus on synthesizing and analyzing the literature that uses AI techniques to fuse multimodal medical data for different clinical applications. \color{black} More specifically, we focus on studies that only fused EHR with medical imaging data to develop various AI methods for clinical applications. \color{black} We present a comprehensive analysis of the various fusion strategies, the diseases and clinical outcomes for which multimodal fusion was used, the ML algorithms used to perform multimodal fusion for each clinical application, and the available multimodal medical datasets.  We followed the PRISMA-ScR (Preferred Reporting Items for Systematic Reviews and Meta-Analyses Extension for Scoping Reviews) guidelines. We searched Embase, PubMed, Scopus, and Google Scholar to retrieve relevant studies. After pre-processing and screening, we extracted data from 34 studies that fulfilled the inclusion criteria. We found that studies fusing imaging data with EHR are increasing and doubling from 2020 to 2021. In our analysis, a typical workflow was observed: feeding raw data, fusing different data modalities by applying conventional machine learning (ML) or deep learning (DL) algorithms, and finally, evaluating the multimodal fusion through clinical outcome predictions. Specifically, early fusion was the most used technique in most applications for multimodal learning (22 out of 34 studies). We found that multimodality fusion models outperformed traditional single-modality models for the same task. Disease diagnosis and prediction were the most common clinical outcomes (reported in 20 and 10 studies, respectively) from a clinical outcome perspective. Neurological disorders were the dominant category (16 studies). From an AI perspective, conventional ML models were the most used (19 studies), followed by DL models (16 studies). Multimodal data used in the included studies were mostly from private repositories (21 studies). Through this scoping review, we offer new insights for researchers interested in knowing the current state of knowledge within this research field.
\end{abstract}
\begin{document}

\flushbottom
\maketitle
%
%
\thispagestyle{empty}


\section*{Introduction}

Over the past decade, digitization of health data have grown tremendously with increasing data repositories spanning the healthcare sectors \cite{ref1}. Healthcare data are inherently multimodal, including electronic health records (EHR), medical imaging, multi-omics, and environmental data. In many applications of medicine, the integration (fusion) of different data sources has become necessary for effective prediction, diagnosis, treatment, and planning decisions by combining the complementary power of different modalities, thereby bringing us closer to the goal of precision medicine\cite{ref2, ref3}.  

Data fusion is the process of combining several data modalities, each providing different viewpoints on a common phenomenon to solve an inference problem. The purpose of fusion techniques is to effectively take advantage of cooperative and complementary features of different modalities \cite{ref4, ref5}. For example, in interpreting medical images, clinical data is often necessary for making effective diagnostic decisions. Many studies found that missing pertinent clinical and laboratory data during image interpretation decreases the radiologists' ability to accurately make diagnostic decisions\cite{ref6}. The significance of clinical data to support the accurate interpretation of imaging data is well established in radiology as well as in a wide variety of imaging-based medical specialties such as dermatology, ophthalmology, and pathology that depend on clinical context to interpret imaging data correctly\cite{ref7,ref8, ref9}. 

Thanks to the advances of AI and ML models, one can achieve a useful fusion of multimodal data with high-dimensionality \cite{ref10}, various statistical properties, and different missing value patterns\cite{ref11}. Multimodal ML is the domain that can integrate different data modalities. In recent years, multimodal data fusion has gained much attention for automating clinical outcome prediction and diagnosis. This can be seen in Alzheimer’s disease diagnosis and prediction \cite{ref12, ref13, ref14, ref15} when imaging data were combined with specific lab test results and demographic data as inputs to ML models, and better performance was achieved than the single-source models. Similarly, fusing pathological images with patient demographic data observed an increase in performance in comparison with single modality models for breast cancer diagnosis\cite{ref16}. Several studies found similar advantages in various medical imaging applications, including diabetic retinopathy prediction, COVID-19 detection, and glaucoma diagnosis\cite{ref17,ref18,ref19}.

This scoping review focuses on studies that use AI models to fuse medical images with EHR data for different clinical applications. Modality fusion strategies play a significant role in these studies. In the literature, some other reviews have been published on the use of AI for multimodal medical data fusion \cite{ref71, ref72, ref73, ref74,ref75,ref29,ref21}; however, they differ from our review in terms of their scope and coverage. Some previous studies focused on the fusion of different medical imaging modalities \cite{ref71, ref72}; they did not consider the EHR in conjunction with imaging modalities. Other reviews focused on the fusion of omics data with other data modalities using DL models \cite{ref73,ref74}. Another study \cite{ref75} focused on the fusion of various internet of medical things (IoMTs) data for smart healthcare applications. Liu et al. \cite{ref20} focused exclusively on integrating multimodal EHR data, where multimodality refers to structured data and unstructured free texts in EHR, using conventional ML and DL techniques. Huang et al. \cite{ref21} discussed fusion strategies of structured EHR data and medical imaging using DL models emphasizing fusion techniques and feature extraction methods. Furthermore, their review covered the research till 2019 and retrieved only 17 studies. In contrast, our review focuses on studies using conventional ML or DL techniques with EHR and medical imaging data, covering 34 recent studies. Table \ref{tab:6} provides a detailed comparison of our review with existing reviews.

\color{black}
\begin{table}
\centering

\begin{tabular}{|p{1.5in}|p{0.2in}|p{2.4in}|p{2.4in}|} \hline 

\textbf{Previous Reviews} & \textbf{Year} & \textbf{Scope and Coverage} & \textbf{Comparative contribution of our review} \\ \hline 
A review on multimodal medical image fusion: Compendious analysis of medical modalities, multimodal databases, fusion techniques and quality metrics \cite{ref71} & 2022 & Their review focused on the fusion of different medical imaging modalities.\newline  &Our review focused on the fusion of medical imaging with multimodal EHR data and considered different imaging modalities as a single modality.  The two reviews did not share any common studies.\newline  \\ \hline 
Advances in multimodality data fusion in neuroimaging \cite{ref72} \newline \newline ~\newline  & 2021 &Their review focused on the fusion of different imaging modalities, considering neuroimaging applications for brain diseases and neurological disorders.   & Our review focused on the fusion of medical imaging with EHR data, considering various diseases, such as neurological disorders, cancer, cardiovascular diseases, psychiatric disorders, eye diseases, and Covid-19. The two reviews did not share any common studies. \\ \hline 
An overview of deep learning methods for multimodal medical data mining \cite{ref73} \newline  & 2022 & Their review focused on the fusion of different types of multi-omics data with EHR and different imaging modalities, only considering DL models for specific diseases (COVID-19, cancer, and Alzheimer's).  & Our review focused on the fusion of medical imaging with EHR data, considering all AI models for various diseases, such as neurological disorders, cancer, cardiovascular diseases, psychiatric disorders, eye diseases, and Covid-19. The two reviews did not share any common studies. \\ \hline 
Multimodal deep learning for biomedical data fusion: a review \cite{ref74} & 2022 & Their review focused on the fusion of different types of multi-omics data with EHR and imaging modalities, considering only DL models. Moreover, they did not provide a summary of the freely accessible multimodal datasets and a summary of evaluation measures used to evaluate the multimodal models.  & Our review focused on the fusion of medical imaging with EHR data, considering all AI models. Moreover, o  ur study provided a summary of the accessible multimodal datasets and a summary of evaluation measures used to evaluate the multimodal models. The two reviews only shared two common studies.  \textbf{} \\ \hline 
A comprehensive survey on multimodal medical signals fusion for smart healthcare systems \cite{ref75}  & 2021 & Their survey did not focus on fusing medical imaging with EHR but rather covered the fusion of IoMTs data for smart healthcare applications and covered studies published untill 2020. Moreover, in their review, multimodality referred to fusing either different 1D medical signals (such as electrocardiogram (ECG) and biosignals), different medical imaging modalities, or 1D medical signals with imaging. & Our review focused on the fusion of medical imaging with EHR (structured and unstructured) for different clinical applications. It included 34 studies, most of them published in 2021 and 2022, with no study common between the two reviews. \\ \hline 
Machine learning for multimodal electronic health records-based research: Challenges and perspectives \cite{ref20} & 2021 & Their review focused on the fusion of structured and unstructured EHR data and did not consider medical imaging modalities. Moreover, they did not provide a summary of the freely accessible multimodal datasets and a summary of evaluation measures used to evaluate the multimodal models.  & Our review focused on the fusion of medical imaging with EHR and considered structured and unstructured data in EHR as a single modality.  The two reviews did not share any common studies.  \\ \hline 
Fusion of medical imaging and electronic health records using deep learning: a systematic review and implementation guidelines \cite{ref21}  & 2020 & Their review focused on the fusion of structured EHR data and medical imaging, considering only DL models, and included only 17 studies published until 2019. & Our review focused on the fusion of medical imaging with EHR data, considering all AI models, and included 34 studies, almost more than half published in 2020 and 2021.  \\ \hline 
\end{tabular}
 \caption{\label{tab:6} Comparison with previous reviews.}
\end{table}

\color{black}

The primary purpose of our scoping review is to explore and analyze published scientific literature that fuses EHR and medical imaging using AI models. Therefore, our study aims to answer the following questions:
\color{black}
\begin{enumerate}
  \item Fusion Strategies: what fusion strategies have been used by researchers to combine medical imaging data with EHR? What is the most used method? 

  \item Diseases: For what type of diseases are fusion methods implemented? 
   \item Clinical outcomes and ML methods: What types of clinical outcomes are addressed using the different fusion strategies? What kind of ML algorithms are used for each clinical outcome? 
  \item Resource: What are the publicly accessible medical multimodal datasets? 
\end{enumerate}

We believe that this review will provide a comprehensive overview to the readers on the advancements made in multimodal ML for EHRs and medical imaging data. Furthermore, the reader will develop an understanding of how ML models could be designed to align data from different modalities for various clinical tasks. Besides, we believe that our review will help identify the lack of multimodal data resources for medical imaging and EHR, thus motivating the research community to develop more multimodal medical data.

\section*{Preliminaries} \label{sec:pre}
We first identify the EHR and medical imaging modalities that are the focus of this review. Then, we present the data fusion strategies that we use to investigate the studies from the perspective of multimodal fusion. 

\subsection*{Data modalities}
In this review, we focus on studies that use two primary data modalities:
\begin{itemize}
    \item Medical imaging modality: This includes N-dimensional imaging information acquired in clinical practice, such as X-ray, Magnetic Resonance Imaging (MRI), functional MRI (fMRI), structural MRI (sMRI), Positron Emission Tomography (PET), Computed Tomography (CT), and Ultrasound. 
    \item EHR data: This includes both structured and unstructured free-text data. Structured data include coded data such as diagnosis codes, procedure codes, numerical data such as laboratory test results, and categorical data such as demographic information, family history, vital signs, and medications. Unstructured data include medical reports and clinical notes.
\end{itemize}
In our review, we consider studies combining the two modalities of EHR and imaging. However, there exist cases where the data could contain only multiple EHR modalities (structured and unstructured) or multiple imaging modalities (e.g., PET and MRI). We consider such data as a single modality, i.e., the EHR modality or imaging modality.

\subsection*{Fusion strategies}
As outlined in \cite{ref21}, fusion approaches can be categorized into early, late, and joint fusion. These strategies are classified depending on the stage in which the features are fused in the ML model. Our scoping review follows the definitions in \cite{ref21} and attempts to match each study to its taxonomy. In this section, we briefly describe each fusion strategy: 

\begin{itemize}
    \item Early fusion: It joins features of multiple input modalities at the input level before being fed into a single ML algorithm for training\cite{ref21}. The modality features are extracted either manually or by using different methods such as neural networks (NN), software, statistical methods, and word embedding models. When NN are used to extract features, early fusion requires training multiple models: the feature extraction models and the single fusion model. There are two types of joint fusion: type I and type II. Type I fuses the original features without extracting features, while type II fuses extracted features from modalities. 
    \item Late fusion: It trains separate ML models on data of each modality, and the final decision leverages the predictions of each model\cite{ref21}. Aggregation methods such as weighted average voting, majority voting, or a meta-classifier are used to make the final decision. This type of fusion is often known as decision-level fusion. 
    \item Joint fusion: It combines the learned features from intermediate layers of NN with features from other modalities as inputs to a final model during training\cite{ref21}. In contrast to early fusion, the loss from the final model is propagated back to the feature extraction model during training so that the learned feature representations are improved through iterative updating of the feature weights. NNs are used for joint fusion since they can propagate loss from the final model to the feature extractor(s). There are two types of joint fusion: type I and type II. The former is when NNs are used to extract features from all modalities. The latter is when not all the input modalities’ features are extracted using NNs\cite{ref21}.
\end{itemize}

\section*{Methods} \label{sec:methods}
In this scoping review, we followed the guidelines recommended by the PRISMA-ScR \cite{ref22}.

\subsection*{Search strategy}
In a structured search, we searched four databases, including Scopus, PubMed, Embase, and Google Scholar, to retrieve the relevant studies. We note here that MEDLINE is covered in PubMed \color{black}. For Google Scholar search results, we selected the first 110 relevant studies, as, beyond 110 entries, the search results rapidly lost relevancy and were unmatched to our review's topic. \color{black} Furthermore, we limited our search to English-language articles published in the last seven years between January 1, 2015, and January 6, 2022. The search was based on abstracts and titles and was conducted between January 3 and January 6, 2022.

In this scoping review, we focused on applying AI models to multimodal medical data-based applications. The term multimodal refers to combining medical imaging and EHR, as described in \hyperref[sec:pre]{Preliminaries} section. Therefore, our search string incorporated three major terms connected by AND:( (“Artificial Intelligence” OR “machine learning” OR “deep learning”) AND “multimodality fusion” AND (“medical imaging” OR “electronic health records”)). We used different forms of each term. We provide the complete search string for all databases in Appendix 1 of the supplementary material. 

\subsection*{Inclusion and exclusion criteria }
We included all studies that fused EHR with medical imaging modalities using an AI model for any clinical application. As AI models, we considered classical ML models, DL models, transfer learning, ensemble learning, etc as mentioned in the search terms in Appendix 1 of the supplementary material. We did not consider studies that use classical statistical models such as regression in our review.  \color{black} Our definition of imaging modalities is any type of medical imaging used in clinical practice, such as MRI, PET, CT scans, and Ultrasound. We considered both structured and unstructured free-text patients’ data for EHR modalities as described in \hyperref[sec:pre]{Preliminaries} section . Only peer-reviewed studies and conference proceedings were included. Moreover, all included studies were limited to English language only. We did not enforce restrictions on types of disorders, diseases or clinical tasks.

We excluded studies that used a single data modality. Also, we excluded studies that used different types of data from the same modality, such as studies that only combined two or more imaging types (e.g. PET and MRI), as we considered this single modality. Moreover, studies that integrated original imaging modalities with extracted imaging features were excluded as this was still considered a single modality. Also, studies that combined multi-omics data modality were excluded. In addition, studies that were unrelated to the medical field or did not use AI-based models were excluded. We excluded reviews, conference abstracts, proposals, editorials, commentaries, letters to editors, preprints, and short letters articles.  Non-English publications were also excluded.   

\subsection*{Study selection}
We used Rayyan web-based review management tool \cite{ref23} for the first screening and study selection. After removing duplicates, we screened the studies based on title and abstract. Subsequently, full-text of the selected studies from the title and abstract screening were assessed for eligibility using our inclusion and exclusion criteria. Two authors (F.M. and H.A.) conducted the study selection and resolved any conflict through discussion. A third author (Z.S.) was consulted when an agreement could not be reached.

\subsection*{Data extraction}
From the final included studies, a data extraction form was designed and piloted on four studies to develop a systematic and accurate data extraction process. The extracted data from the studies are first author’s name, year, the country of the first author's institution, \color{black} disease's name, clinical outcome, imaging modality, EHR modality, fusion strategy, feature extraction methods, data source, AI models, evaluation metrics, and comparison with single modality. In Appendix 2 of the supplementary material, we provide the extracted information description in detail. One author (F.M.) performed the data extraction, and two other authors (Z.S. and H.A.) reviewed and verified the extracted data. Any disagreement was resolved through discussion and consensus between the three authors.\color{black}

\subsection*{Data synthesis}
Following the data extraction, we used a narrative approach to synthesize the data. We analyzed the studies from five perspectives: fusion strategies, diseases, clinical outcomes with ML algorithms, data sources/type, and evaluation mechanism. For fusion strategies, we focused on how the multimodal data was fused. In addition, we recorded implementation details of the model, such as feature extraction and single modality evaluation. We also extracted information on the diseases for which fusion methods were implemented. Furthermore, we analyzed where the data fusion models were applied for clinical outcomes and what ML models were used for each task. Moreover, we focused on the type of imaging and EHR data used by the studies, the source of data, and its availability. Finally, for evaluation, we focused on the evaluation metrics used by each study. 

\subsection*{Study quality assessment}
In accordance with the guidelines for scoping reviews \cite{ref24,ref25}, we did not perform quality assessments of the included studies. 

\section*{Results}
\subsection*{Search results}
\begin{figure}
\centering
\includegraphics[width=\linewidth]{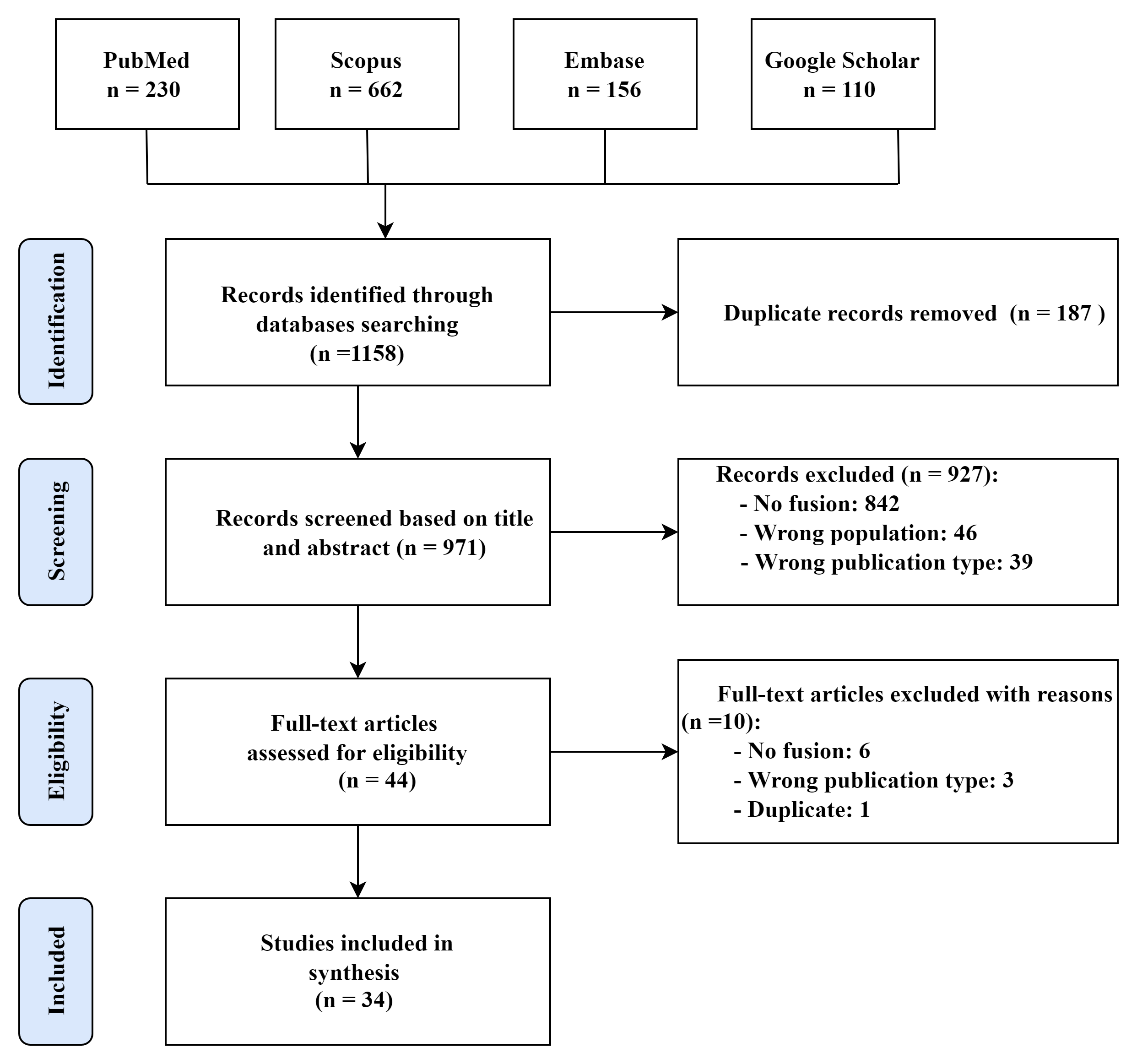}
\caption{PRISMA flow chart for study identification, screening, and selection.}
\label{fig:1}
\end{figure}

A total of 1158 studies were retrieved from the initial search. After duplicates elimination, 971 studies were retained. Based on our study selection criteria (see \hyperref[sec:methods]{Methods}), 44 studies remained for full-text review after excluding articles based on their abstract and title. Moreover, 10 studies were removed after the full-text screening. Finally, 34 studies met our inclusion criteria and were selected for data extraction and synthesis. Figure \ref{fig:1} shows a flowchart of the study screening and selection process.

\subsection*{Demographics of the studies}
\begin{figure}[ht]
\centering
\includegraphics [width=\textwidth]{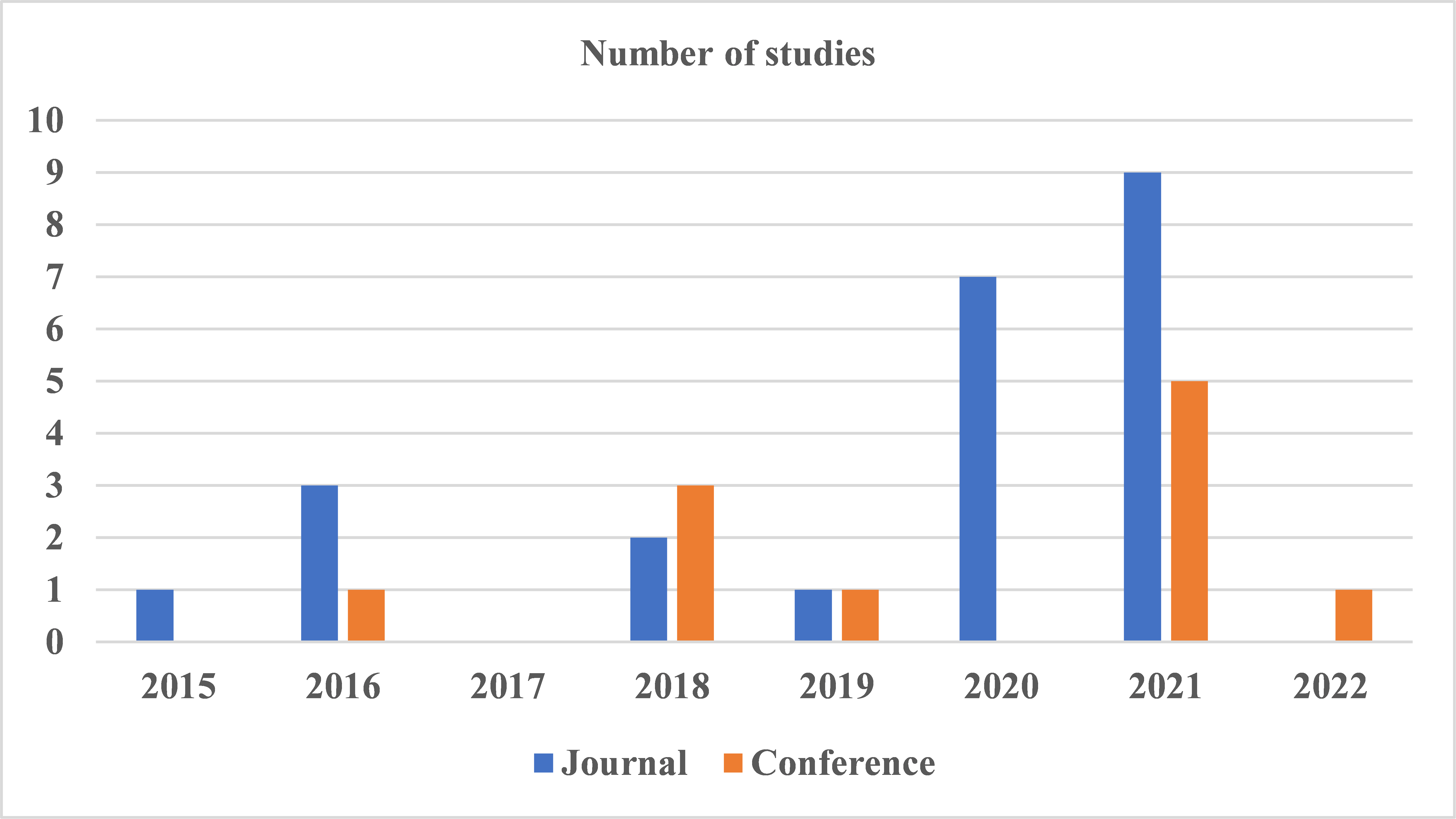}
\caption{The distribution of studies by the   type of publication and the year.}
\label{fig:2}
\end{figure}

As presented in Table \ref{tab:1},  approximately two-thirds of the studies were journal articles ($n$= $23$, $\sim 68$\%)\cite{ref12, ref13,ref14,ref15, ref17,ref19, ref26,ref27,ref28,ref29,ref30,ref31,ref32,ref33,ref34,ref35,ref36,ref37,ref38,ref39,ref40,ref41}, whereas $11$ studies were conference proceedings {($\sim 32$\%)} \cite{ref16, ref42,ref43,ref44,ref45,ref46,ref47,ref48,ref49,ref50,ref51}. \color{black} Most of the studies were published between 2020 and 2022 ($n$ = $22$, $\sim~ 65\%$). Figure \ref{fig:2} shows a visualization of the publication type-wise and year-wise distribution of the studies. The included studies were published in $13$ countries; however, most of these studies were from the USA ($n$ = $10$, $\sim30\%$) and China ($n$ = $8$, $\sim 24\%$).
\begin{table}[ht]
\centering
\begin{tabular}{|l|l|}
\hline
\textbf{Characteristics} & \textbf{Number of studies}  \\
\hline
\textbf{Year} &   \\
\hline
\hspace{0.3cm}2022 & 1  \\
\hline
\hspace{0.3cm}2021 & 14  \\
\hline
\hspace{0.3cm}2020 & 7  \\
\hline
\hspace{0.3cm}2019 & 2  \\
\hline
\hspace{0.3cm}2018 & 5  \\
\hline
\hspace{0.3cm}2016 & 4  \\
\hline
\hspace{0.3cm}2015 & 1  \\
\hline

\textbf{Country} &   \\
\hline
\hspace{0.3cm}United States of America (USA)	& 10 \\
\hline
\hspace{0.3cm}China &	8 \\
\hline
\hspace{0.3cm}United Kingdom (UK)	& 4\\
\hline
\hspace{0.3cm}Germany & 2 \\
\hline
\hspace{0.3cm}India &	2 \\
\hline
\hspace{0.3cm}Australia &	1 \\
\hline
\hspace{0.3cm}Denmark	 & 1 \\
\hline
\hspace{0.3cm}Iran	& 1 \\
\hline
\hspace{0.3cm}Korea	& 1 \\
\hline
\hspace{0.3cm}Pakistan &	1\\
\hline
\hspace{0.3cm}Kingdom of Saudi Arabia &	1\\
\hline
\hspace{0.3cm}Singapore &	1 \\
\hline
\textbf{Publication type} & \\
\hline
\hspace{0.3cm}Journal & 23\\
\hline
\hspace{0.3cm}Conference & 11\\
\hline

\end{tabular}
\caption{\label{tab:1}Demographics of the included studies.}
\end{table}

\subsection*{Data fusion strategies}
 We mapped the included studies to the taxonomy of fusion strategies outlined in the \hyperref[sec:pre]{Preliminaries} Section. A primary interest of our review is to identify the fusion strategies that the included studies used to improve the performance of ML models for different clinical outcomes. 

\subsubsection*{Early fusion}
The majority of the included studies ($n$ = $22$, $\sim 65 \%$) used early fusion to combine medical imaging and non-imaging data. When the input modalities have different dimensions, such as when combining one-dimensional (1D) EHR data with 2D or 3D imaging data, it is essential to extract high-level imaging features in 1D before fusing with 1D EHR data. To accomplish this, various methods were used in the studies, including neural network-based features extraction, data generation through software, or manual extraction of features. Out of the $22$ early fusion studies, $19$ studies \cite{ref12,ref13,ref15,ref27,ref28,ref29,ref30,ref31,ref36,ref37,ref38,ref39,ref40,ref34,ref45,ref46,ref47,ref48} used manual or software-based imaging features, and $3$ studies used neural network-based architectures to extract imaging features before combining with other clinical data modality \cite{ref16,ref18,ref49}. Six out of the $19$ studies that used manual or software-based features reduced the feature dimension before concatenating the two modalities’ features using different methods \cite{ref29,ref31,ref40,ref45, ref46,ref47}. Such methods include recursive feature elimination \cite{ref47}, a filter-based method using Pearson correlation coefficient\cite{ref46}, Random Forest feature selection based on Gini importance\cite{ref45}, Relief-based feature selection method\cite{ref29}, a wrapper-based method using backward feature elimination\cite{ref31}, and a rank-based method using Gini coefficients\cite{ref40}. Moreover, $3$ studies \cite{ref13,ref15,ref39} utilized the principal component analysis (PCA) dimensionality reduction technique to reduce the feature dimension.

In the studies that used neural network-based architectures to extract imaging features, CNN  architectures were used in three studies \cite{ref16,ref18,ref49}. These studies concatenated the multimodal features (CNN-extracted and EHR features) for their fusion strategy. 

Fourteen early fusion studies evaluated their fusion models' performance against that of single modality models \cite{ref12,ref13,ref15,ref16,ref18,ref26, ref27,ref28,ref29,ref31,ref36, ref37, ref38,ref39,ref46} \sloppy. As a result, $13$ of these studies exhibited a better performance for fusion when compared with their imaging-only and clinical-only counterparts \cite{ ref12,ref13,ref15,ref16,ref18,ref26, ref27,ref28,ref29,ref36, ref37, ref38,ref39,ref46}.
\subsubsection*{Joint fusion}
Joint fusion was the second most common fusion strategy used in $10$ out of the $34$ studies. In these studies, different neural network-based methods were used for processing the imaging and EHR data modalities. Chen et al.\cite{ref34} used the Visual Geometry Group  (VGG-16) architecture to extract features from MRI images, while they used a bidirectional long-short term memory (LSTM ) network with an attention layer to learn feature representation from MRI reports. Then, they concatenated the learned features of the two modalities before feeding them into a stacked K-nearest neighbor (KNN) attention pooling layer. Grant et al.\cite{ref50} used a Residual Network (ResNet50) architecture to extract relevant features from the imaging modality and fully connected NN to process the non-imaging data. They directly concatenated the learned feature representation of the imaging and non-imaging data and fed them into two fully connected networks. Yidong et al.\cite{ref19} used a Bayesian CNN encoder-decoder to extract imaging features and a Bayesian Multilayer perception (MLP) encoder-decoder to process the medical indicators data. The study directly concatenated the two feature vectors and fed the resulted vector into another Bayesian MLP. Samak et al.\cite{ref42} utilized CNN with a self-attention mechanism to extract the imaging features and fully connected NNs to process the metadata information. Lili et al. \cite{ref34} used VGG-19 architecture to extract the multimodal MRI features and fully connected networks for clinical data. The study concatenated the two feature vectors and fed them into fully connected NN. Another study\cite{ref41} applied CNN layers for imaging features extraction and word embeddings (Word2vec) with self-attention for textual medical data. In another research \cite{ref33}, Fang et al. applied a ResNet architecture and MLP for imaging and clinical data feature extraction. Then, the authors fused the feature vectors by concatenation and fed them into an LSTM network followed by a fully connected network. Hsu et al.\cite{ref17} concatenated the imaging features extracted using Inception-V3 model with the clinical data features before feeding them to fully connected NN. In\cite{ref51}, Sharma et al. used CNN to extract image features and then concatenated them directly with the clinical data to feed into a SoftMax classifier. Xu et al.\cite{ref48} used AlexNet architecture to convert the imaging data into a feature vector fusible with other non-image modalities. Then, they jointly learned the non-linear correlations among all modalities using fully connected NN. Out of $10$ joint fusion studies, seven studies evaluated their fusion models’ performance against that of a single modality and reported a performance improvement when fusion was used \cite{ref17,ref34,ref41,ref42,ref44,ref48,ref50}.

\subsubsection*{Late fusion}
Late fusion was the least common fusion approach used in the included studies, as only two studies used it. Qiu et al.\cite{ref32} trained three independent imaging models that took a single MRI slice as input, then aggregated the prediction of these models using maximum, mean, and majority voting. After combining the results of these aggregations by majority vote, the study performed late fusion with the clinical data models. In another study \cite{ref35}, Huang et al. trained four different late fusion models. Three models took the average of the predicted probabilities from the imaging and EHR modality models as the final prediction. The fourth model used an NN classifier as an aggregator, which took as input the single modality models’ prediction. The study also created early, joint fusion models and two single modality models to compare with late fusion performance. As a result, the late fusion outperformed both the early and joint fusion models and the single modality models.

\subsection*{Diseases}
We categorized the diseases and disorders in the included studies into seven types: neurological disorders, cancer, cardiovascular diseases, Covid-19, psychiatric disorders, eye diseases, and other diseases. The majority of the included studies focused on neurological disorders ($n$ = $16$). Table \ref{tab:2} shows the distribution of the included studies in terms of the diseases and disorders they covered. 
\begin{table}[ht]
\centering
\begin{tabular}{|l|l|l|}\hline

\textbf{Disease Category} & \textbf{Number of studies} & \textbf{Study reference} \\ \hline 
\textbf{Neurological disorders} & \textbf{18} & \textbf{\textit{}} \\ \hline 
\hspace{0.3cm} Alzheimer's disease (AD)\textbf{} & 7 & \cite{ref12,ref13,ref14,ref15,ref39,ref43,ref44}\textbf{} \\ \hline 
\hspace{0.3cm} Mild cognitive impairment (MCI) & 4 & \cite{ref32,ref37,ref45,ref46} \\ \hline 
\hspace{0.3cm} Ischemic Stroke  & 2 & \cite{ref30,ref42}  \\ \hline 
\hspace{0.3cm} Demyelinating diseases & 1 & \cite{ref26} \\ \hline 
\hspace{0.3cm} Neurodevelopmental Deficits & 1 & \cite{ref34} \\ \hline 
\hspace{0.3cm} Epilepsy\textbf{} & 1 & \cite{ref28} \\ \hline 
\textbf{Cancer} & \textbf{5} & \textbf{\textit{}} \\ \hline 
\hspace{0.3cm}Breast Cancer & 2 & \cite{ref16,ref36} \\ \hline 
\hspace{0.3cm}Glioblastoma  & 1 & \cite{ref38} \\ \hline 
\hspace{0.3cm}Lung Cancer & 1 & \cite{ref50} \\ \hline 
\hspace{0.3cm}Upper Gastrointestinal (UGI) Cancer & 1 & \cite{ref41} \\ \hline 
\textbf{Cardiovascular diseases} & \textbf{3} & \textbf{\textit{}} \\ \hline 
\hspace{0.3cm}Aortic stenosis  & 1 & \cite{ref49} \\ \hline 
\hspace{0.3cm}Cardiomegaly  & 1 & \cite{ref50} \\ \hline 
\hspace{0.3cm}Myocardial Infarction & 1 & \cite{ref51} \\ \hline 

\textbf{Psychiatric disorder} & \textbf{2} & \textbf{\textit{}} \\ \hline 
\hspace{0.3cm}Bipolar disorder  & 1 & \cite{ref27} \\ \hline 
\hspace{0.3cm}Schizophrenia & 1 & \cite{ref31} \\ \hline 

\textbf{Eye diseases} & \textbf{2} & \textbf{\textit{}} \\ \hline 
\hspace{0.3cm}Diabetic Retinopathy (DR) & 1 & \cite{ref17} \\ \hline 
\hspace{0.3cm}Glaucoma & 1 & \cite{ref19} \\ \hline 
\textbf{COVID-19} & \textbf{3} & \cite{ref18,ref29,ref33} \\ \hline 
\textbf{Other diseases} & \textbf{3} & \textbf{\textit{}} \\ \hline 
\hspace{0.3cm}Cervical dysplasia & 1 & \cite{ref48} \\ \hline 
\hspace{0.3cm}Pulmonary Embolism (PE) & 1 & \cite{ref35} \\ \hline 
\hspace{0.3cm}Hepatitis B & 1 & \cite{ref47} \\ \hline 
\end{tabular}
\caption{\label{tab:2}Disease distribution covered by the 34 studies.}
\end{table}

\subsection*{Clinical outcomes and machine learning models}
\begin{figure}
\centering
\includegraphics [width=\textwidth]{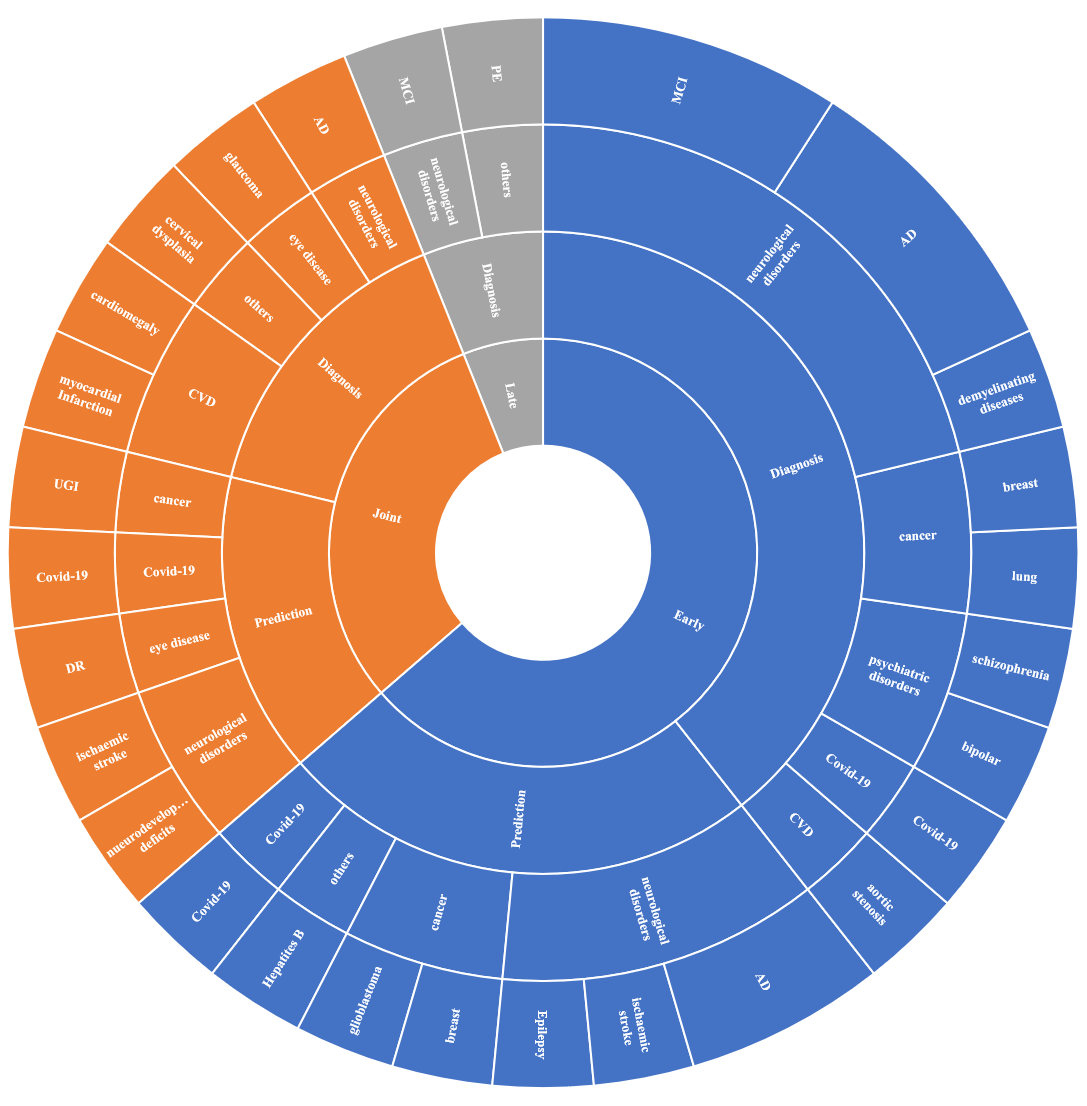}
\caption{Fusion strategies associated with clinical outcomes for different diseases.}
\label{fig:4}
\end{figure}

Multimodal ML enables a wide range of clinical applications such as diagnosis, early prediction, patient stratification, phenotyping, biomarkers identification, etc. In this review, we labeled each study according to its clinical outcome. We categorized the retrieved clinical tasks into two main categories: diagnosis and prediction. Though some of the studies mentioned detection, classification, diagnosis, and prediction, we categorized them under the diagnosis category. Under the early prediction group, we considered only the studies that predict diseases before onset, identify significant risk factors, predict mortality and overall survival, and predict a treatment outcome. These clinical outcomes were implemented using multimodal ML models. This section summarizes the different clinical tasks of the retrieved studies, the fusion strategy used, and the ML models that were developed for each task. Figure \ref{fig:4} shows the distribution of fusion strategies associated with different diseases’ and clinical outcomes.  

\subsubsection*{Diagnosis}
The most common applied clinical outcome in the included studies was the diagnosis, reported in $20$ ($\sim59\%$) studies. In these studies, EHRs were combined with medical imaging to diagnose a spectrum of diseases including neurological disorders ($n$ = $9$) \cite{ref13, ref14,ref15,ref26,ref32,ref37,ref44,ref45,ref4}, psychiatric disorders ($n$ = $2$) \cite{ref27, ref31},  CVD ($n$ =$3$) \cite{ref49, ref50, ref51}, Cancer ($n$ = $2$) \cite{ref16,ref50}, and four studies for other different diseases \cite{ref18,ref19,ref35,ref48}. Specifically, most of the studies that focused on detecting neurological diseases were for AD ($n=4$) \cite{ref13,ref14,ref15,ref44}, and MCI ($n$ = $4$) \cite{ref32,ref37,ref45,ref46}.

Early fusion was the most utilized technique for diagnosis purposes used in $13$ studies. These studies employed different ML models on the fused imaging and EHR data for diagnosing different diseases. Most of these studies were for diagnosing neurological and and psychiatric disorders such as AD \cite{ref13, ref14, ref15}, MCI \cite{ref37,ref45,ref46}, demyelinating diseases \cite{ref26}, bipolar disorder \cite{ref27}, and schizophrenia \cite{ref31} . Parvathy et al. \cite{ref13} reported diagnosing AD by fusing sMRI and PET imaging features with mini-mental state examination (MMSE) score, clinical dementia rating (CDR), and age of the subjects. They fed the fused features vector to different ML models, including support vector machine (SVM), random forest (RF), and gaussian process (GP) for classification. Niyas et al. \cite{ref14} classified AD by fusing MRI, PET, demographic data, and lab tests, including cognitive tests and Cerebro-Spinal Fluid (CSF) test. They applied dynamic ensemble of classifiers selection algorithms using a different pool of classifiers on the fused features for classification. Hamid et al.\cite{ref15} combined MRI and PET imaging features with personal information and neurological data such as MMSE and CRF features for  AD early diagnosis. In their study, they fed the fused features into SVM for classification. For MCI diagnosis, Matteo et al. \cite{ref37} proposed combining MRI imaging with cognitive assessments for MCI diagnosis. They concatenated the features of both modalities and fed them into a linear and quadratic discriminant analysis algorithm for diagnosis. Parisa et al. \cite{ref45,ref46} integrated features extracted from MRI and PET images with neuropsychological tests and demographic data (gender, age, and education) to diagnose MCI early. They trained SVM and deep NNs using the fused features for classification in \cite{ref45} and \cite{ref46}, respectively. In another study \cite{ref26}, Xin et al. combined MRI imaging with structured data extracted from EHRs to diagnose demyelinating diseases using SVM. For bipolar disorder, Rashmin et al. \cite{ref27} combined multimodal imaging features with neuropsychological tests and personal information features. They fed them into SVM to differentiate bipolar patients from healthy patients. Ebdrup et al. \cite{ref31} proposed integrating MRI and diffusion tensor imaging tractography (DTI) imaging with neurocognitive tests and clinical data for schizophrenia classification. Then, they fused the features of the two modalities and fed them to different types of ML classifiers, including SVM, RF, linear regression (LR), decision tree (DT), and Naïve Bayes (NB) for classification.

Moreover, two studies implemented multimodality early fusion to diagnose different cancer diseases \cite{ref16,ref50}. Yan et al. \cite{ref16} fused pathological images and structured data extracted from EHRs to classify malignant and benign breast cancer. Then, they fused the features of the two modalities and fed them to two fully connected NN followed by a SoftMax layer for classification. Seung et al. \cite{ref50} combined PET imaging with clinical and demographic data for differentiating lung adenocarcinoma (ADC) from squamous cell carcinoma. They fed the integrated features into different algorithms such as SVM, RF, LR, NB, and artificial neural network (ANN) for classification. For COVID-19 diagnosis, Ming et al. \cite{ref18} combined CT images with clinical features and fed them into different ML models, including SVM, RF, and KNN for diagnosis. Finally, Tanveer et al.\cite{ref49} combined features from echocardiogram reports and images, with diagnosis information for the detection of patients  with aortic stenosis CVD. Their study fed the combined features to an RF learning framework to detect patients likely to have the disease.

Joint fusion was used for diagnostic purposes in $5$ studies \cite{ref19,ref44,ref48,ref50,ref51}. These studies employed different types of DL architectures to learn and fuse the imaging and EHR data for diagnosis purposes. In \cite{ref19}, they proposed a Bayesian deep multisource learning (BDMSL) model that integrated retinal images with medical indicators data to diagnose glaucoma. For this model, they used Bayesian CNN encoder-decoder to extract imaging features and a Bayesian MLP encoder-decoder to process the medical indicators data. The two feature vectors were directly concatenated and fed into Bayesian MLP for classification. Chen et al.\cite{ref44} used DL for multimodal feature extraction and classification to detect AD; the authors used the VGG-16 model to extract features from MRI images and a bidirectional LSTM network with an attention layer to learn features from MRI reports. Then, they fed the fused features into a stacked KNN pooling layer to classify the patient’s diagnosis data. In \cite{ref48}, Xu et al. proposed an end-to-end deep multimodal framework that can learn better complementary features from the image and non-image modalities for cervical dysplasia diagnosis. They used CNN, specifically AlexNet architecture, to convert the cervigram image data into a feature vector fusible with other non-image modalities. After that, they jointly learned the non-linear correlations among all modalities using fully connected NN for cervical dysplasia classification. Another two studies \cite{ref50,ref51} also employed DL models to jointly learn multimodal feature representation for diagnosing CVDs. The former \cite{ref50} proposed a multimodal network for cardiomegaly classification, which simultaneously integrates the non-imaging intensive care unit (ICU) data (laboratory values, vital sign values, and static patient metadata, including demographics) and the imaging data (chest X-ray). They used a ResNet50 architecture to extract features from the X-ray images and fully connected NN to process the ICU data. To join the learned imaging and non-imaging features, they concatenated the learned feature representation and fed them into two fully connected layers to generate a label for cardiomegaly diagnosis. The latter study \cite{ref51} proposed a stacked multimodal architecture called SM2N2, which integrated clinical information and MRI images. In their research, they used CNN to extract imaging features, and then they concatenated these features with clinical data to feed into a SoftMax classifier for myocardial infarction detection. 

Late fusion was implemented in $2$ studies \cite{ref32,ref35} for disease diagnosis purposes. Fang et al.\cite{ref32} proposed the fusion of MRI scans, logical memory (LM) tests, and MMSE for MCI classification. Their study utilized VGG-11 architecture for MRI feature extraction and developed two MLP models for MMSE and LM test results. Then, they combined both MRI and MLP models using majority voting. As a result, the fusion model outperformed the individual models. Huang et al. \cite{ref35} utilized a non-open dataset comprising CT scans and EHR data to train two unimodal and four late fusion models for PE diagnosis. They used their previously implemented architecture (PENet) \cite{ref52} to encode the CT images and a feedforward network to encode the tabular data. The late fusion approach performed best among the fusion models and outperformed the models trained on the image-only and the tabular-only data.

\subsubsection*{ Early Prediction}
Prediction tasks were reported in $14$ ($\sim41.2\%$) studies. In these studies, EHRs were fused with medical imaging to predict different outcomes, including disease prediction, mortality prediction, survival prediction, and treatment outcome prediction. Ten studies of the prediction tasks were disease prediction \cite{ref12,ref17,ref28,ref33,ref34,ref36,ref39,ref41,ref43,ref47}, which involved determining whether an individual might develop a given disease in the future. The second most common prediction task was treatment outcome prediction reported in $2$ studies \cite{ref30,ref42}, followed by one study for mortality prediction and overall survival prediction \cite{ref29,ref38}, respectively. 

The early fusion technique was used in $6$ studies \cite{ref12,ref28,ref36,ref39,ref43,ref47} for disease prediction. Minhas et al.\cite{ref12} proposed an early fusion model to predict which subjects will progress from MCI to AD in the future. The study concatenated MRI extracted features with demographic and neuropsychological biomarkers before feeding them to an SVM model for prediction. Ali et al. \cite{ref28} proposed a model to predict Epileptogenic-Zone in the Temporal Lobe by feeding MRI extracted features integrated with set-of-semiology features into various ML models such as LR, SVM, and Gradient Boosting. Ma et al.\cite{ref36} fused MRI and clinicopathological features for predicting metachronous distant metastasis (DM) in breast cancer. They fed the concatenated features to an LR model. Another study \cite{ref39} combined MRI-derived features and high-throughput brain phenotyping to diagnose and predict the onset of AD. They fed the fused features into different ML classifiers, including RF, SVM, and LR. Ulyana et al.\cite{ref43} trained a deep, fully connected network as a regressor in a 5-year longitudinal study on AD to predict cognitive test scores at multiple future time points. Their model produced MMSE scores for ten unique future time points at six-month intervals by combing biomarkers from cognitive test scores, PET, and MRI. They early fused imaging features with the cognitive test scores through concatenation before feeding them into the fully connected network. Finally, Bai et al.\cite{ref47} compared different multimodal biomarkers (clinical data, biochemical and hemologic parameters, and ultrasound elastography parameters) for predicting the assessment of fibrosis in chronic hepatitis B using SVM. 

For disease prediction, joint fusion was used in $4$ studies \cite{ref17,ref33,ref34,ref41}. Hsu et al. \cite{ref17} proposed a deep multimodal fusion model that trained heterogeneous data from fundus images and non-image data for DR screening. They concatenated the imaging extracted features from Inception-V3 with the clinical data features before feeding them to fully connected NN followed by SoftMax layer for classification. Fang et al.\cite{ref33} developed a prediction system by jointly fusing CT scans and clinical data to predict the progression of COVID-19 malignancy. In their study, the feature extraction part applied a ResNet architecture and MLP for CT and clinical data, respectively. Then, they concatenated the different features and fed them into an LSTM network followed by a fully connected NN for prediction. In\cite{ref34}, the authors proposed a deep multimodal model for predicting neurodevelopmental deficits at 2 years of age. Their model consisted of a feature extractor and fusion classifier. In the feature extractor, they used VGG-19 architecture to extract MRI features and fully connected NN for clinical data. Then, the study combined the extracted features of the two modalities and fed their combination to another fully connected network in the fusion classifier for prediction. To evaluate the performance of the modality fusion, they tested their model using a single modality of MRI and clinical features. The results showed that multimodal fusion outperformed the single modality performance. Another study \cite{ref41} also used multimodal joint fusion for UGI cancer screening. Their model integrated features extracted from UGI endoscopic images with corresponding textual medical data. They applied CNN for image feature extraction and word embeddings (Word2vec) with self-attention for textual medical data feature extraction. After that, they concatenated the extracted features of the two modalities and fed them into fully connected NN for prediction. Their results showed that multimodal fusion outperformed the single modality performance.

For treatment outcome prediction \cite{ref30,ref42}, the former \cite{ref30} implemented early fusion while the latter \cite{ref42} used joint fusion. For acute ischemic stroke, Gianluca et al. \cite{ref30} evaluated the predictive power of imaging, clinical, and angiographic features to predict the outcome of acute ischemic stroke using ML. The study early fused all features into gradient boosting classifiers for prediction. In \cite{ref42}, the authors proposed a DL model to directly exploit multimodal data (clinical metadata and non-contrast CT (NCCT) imaging data) to predict the success of endovascular treatment for ischemic stroke. They utilized CNN with a self-attention mechanism to extract the features of images, and then they concatenated them with the metadata information. Then, the classification stage of the proposed model processed the fused features through a fully connected NN, followed by the Softmax function applied to the outputs. Their results showed that multimodal fusion outperformed the single modality performance. 

Both the mortality and overall survival prediction studies \cite{ref29,ref38} implemented early fusion. In \cite{ref29}, they developed a model to predict COVID-19 ventilatory support and mortality early on to prioritize patients and manage the hospital resources’ allocation. They fused patients’ CT images and EHR data features by concatenation before feeding them to different ML models, including SVM, RF, LR, and eXtreme gradient boosting. They evaluated the performance against single modality models and observed that the results for multimodal fusion were better. The other study \cite{ref38} aimed to develop ML models to predict glioblastoma patients’ overall survival (OS) and progression‐free survival (PFS) based on combining treatment features, pathological, clinical, PET/CT‐derived information, and semantic MRI‐based features. They concatenated the features of all modalities and fed them to an RF model. The study showed that the model based on multimodal fusion data outperformed the single modality models.

\subsection*{Datasets}
\subsubsection*{Patient Data Types}
The included studies reported medical imaging and EHRs (structured and non-structured) patient’s data types. In terms of imaging modality, CT, MRI, fMRI, structural MRI (sMRI), PET, Diffusion MRI, DTI, ultrasound, X-ray, fundus images, and PET were used in the studies. MRI and PET images were the most utilized modalities. Out of the included $34$ studies, $13$ used MRI images, and $8$ used PET images mostly for AD diagnosis and prediction. In terms of EHRs, structured data was the most commonly used modality ($n$ = $32$). Table \ref{tab:3} summarizes the types of imaging and EHR data used in the studies.

\begin{table}[ht]
\centering
\begin{tabular}{|l|l|l|}\hline
\textbf{Data Type} & \textbf{Number of studies} & \textbf{Study reference} \\ \hline 
\textbf{Imaging Data}   &   & \\ \hline 
      \hspace{0.2 cm}MRI imaging &  &  \\ \hline 
      \hspace{0.5 cm} MRI & $13$ & \cite{ref12,ref14,ref15,ref26,ref27,ref32,ref36,ref37,ref38,ref43,ref44,ref45,ref46}  \\ \hline 
      \hspace{0.5 cm} DTI &  $3$ & \cite{ref27,ref31,ref34} \\ \hline 
      \hspace{0.5 cm} fMRI & $2$ & \cite{ref27,ref34} \\ \hline 
      \hspace{0.5 cm} sMRI and Diffusion MRI &  $1$ & \cite{ref39} \\ \hline 
       \hspace{0.2 cm}PET &  8 & \cite{ref13, ref14,ref15,ref38,ref40,ref43,ref45,ref46} \\ \hline 
       \hspace{0.2 cm}CT & $7$ & \cite{ref18,ref30,ref33,ref35,ref38,ref40,ref42} \\ \hline 
        \hspace{0.2 cm}X-ray & $2$ & \cite{ref29,ref50} \\ \hline 
        \hspace{0.2 cm}fundus images & $2$ & \cite{ref17,ref19} \\ \hline 

       \hspace{0.2 cm}Ultrasound & $1$ & \cite{ref47} \\ \hline 
      \hspace{0.5 cm}Echocardiography  & $1$ & \cite{ref49} \\ \hline 
        \hspace{0.2 cm}Pathological images & $1$ & \cite{ref16} \\ \hline 
        \hspace{0.2 cm}Cervigram images & $1$ & \cite{ref48} \\ \hline 
        \hspace{0.2 cm}Endoscopy images & $1$ & \cite{ref41} \\ \hline 
\textbf{EHR Data } &  &  \\ \hline 
        \hspace{0.2 cm}Structured  & $32$ & \cite{ref12, ref14,ref13, ref15,ref16, ref17,ref18, ref19, ref26, ref27,ref28,ref29,ref30,ref31, ref32,ref33, ref34, ref35,ref36,ref37,ref38,ref39, ref40, ref42,ref43,ref45,ref46, ref47, ref48, ref49,ref50,ref51} \\ \hline 
        \hspace{0.2 cm}Unstructured  & $2$ & \cite{ref41,ref44} \\ \hline 
\end{tabular}

\caption{\label{tab:3}Patient data types used in the included studies.}
\end{table}

\subsubsection*{Patient data Resources}
Almost two-thirds of the studies included in this scoping review used private data sources (clinical data that are not publicly available) ($n$ = $21$, $\sim 59\%$). In contrast, publicly accessible datasets were used in only $13$ studies. We observed that the most used public dataset was the “Alzheimer's Disease Neuroimaging Initiative” dataset (ADNI) \cite{ref53} , where $7$ out of $13$ studies used the dataset. Other publicly available datasets that were used among the included studies were the “National Alzheimer's Coordinating Center” (NACC) dataset \cite{ref54}, the “Medical Information Mart for Intensive Care” (MIMIC-IV) dataset \cite{ref55}, the "National Cancer Institute" (NCI) dataset, ADNI TADPOLE dataset \cite{ref70} , and MR CLEAN Trial dataset \cite{ref56}. In Table \ref{tab:4}, we summarize the public multimodal medical datasets and their clinical applications. Considering these datasets for each clinical task, the most popular is ADNI for AD and MCI disease diagnosis and prediction.

\begin{table}
\centering
\begin{tabular}{|p{2cm}|p{4cm}|p{4cm}|p{4cm}|p{2cm}|}\hline
\textbf{Public Dataset} & \textbf{Description} & \textbf{URL} & \textbf{Clinical outcomes} & \textbf{Study reference} \\ \hline 
ADNI &   ADNI represents a series of studies, including ADNI 1, 2, and 3, designed to study MCI and its progression into AD. It has MRI and PET images along with clinical and genetic information \cite{ref53}. & \url{https://adni.loni.usc.edu/data-samples/data-types/}  & Disease diagnosis (AD)\newline \newline Disease diagnosis (MCI)\newline \newline  Disease Prediction (AD)\newline \newline  & \cite{ref13,ref15} \newline \newline  \cite{ref45,ref46} \newline \newline \cite{ref12,ref39,ref43} \\ \hline 
ADNI TADPOLE  & ADNI has a simplified counterpart, TADPOLE, which has a subset of ADNI-3 samples and features.ATDPOLE does not include raw images, but it has processed structural information about the images such as ROI averages, thicknesses of the cortex and volumes of brain sub-regions, etc  \cite{ref70}.& \url{https://tadpole.grand-challenge.org/Data/} \newline  & Disease diagnosis (AD) & \cite{ref14} \\ \hline 
NACC\textbf{} & The NACC dataset was established to facilitate collaborative AD research. The dataset comprises MRI data, demographic data, neuropsychological testing scores, and clinical diagnosis of patients \cite{ref54}.\newline & \url{ https://naccdata.org/requesting-data/nacc-data} & Disease Diagnosis (MCI) & \cite{ref32} \\ \hline 

MIMIC-CXR, MIMIC-IV \textbf{} & MIMIC-CXR is a dataset of patient chest radiographs. It contains X-ray studies for 64,588 patients \cite{ref57}.\textbf{\newline }MIMIC-IV is a database for patients admitted to critical care units comprising patient stay information, patient's ICU data, and lookup tables to allow linking to MIMIC-CXR \cite{ref55}. & MIMIC-CXR:\url{https://www.nature.com/articles/s41597-019-0322-0} \newline \newline MIMIC-IV:\url{https://physionet.org/content/mimiciv/0.4/} \newline \newline  & Disease diagnosis (Cardiomegaly) & \cite{ref50} \\ \hline 
NCI & Data collections produced by major NCI initiatives are listed in the NCI Data Catalog, including Clinical data, Genomics, imaging, and Proteomics.\newline & \url{ https://datascience.cancer.gov/resources/nci-data-catalog}\textbf{} & Disease diagnosis (Cervical dysplasia) \textbf{} & \cite{ref48} \\ \hline 
MR CLEAN Trial\textbf{} & A longitudinal study of 500 patients treated with endovascular therapy in The Netherlands for acute ischemic stroke comprising NCCT images, CT Angiography (CTA) images, and clinical metadata information on its patients \cite{ref56}. & \url{ https://www.mrclean-trial.org/home.html}& Treatment outcome prediction (ischemic stroke) & \cite{ref42} \\ \hline 
\end{tabular}
\caption{\label{tab:4}Multimodal medical datasets and clinical outcome applications.}
\end{table}

\subsection*{Evaluation metrics}
Evaluation metrics are mainly dependent on the clinical task. Typically, accuracy, the area under the curve (AUC), sensitivity, specificity, F1- measure, and precision are mostly used for the evaluation of diagnosis and prediction tasks. Table \ref{tab:5} shows the distribution of the evaluation measures used in the included studies
\begin{table}[ht]
\centering
\begin{tabular}{|p{6cm}|p{3cm}|p{7cm}|}\hline
\textbf{Evaluation metrics} & \textbf{Number of studies } & \textbf{Study Reference} \\ \hline 
Accuracy &   $31$ & \cite{ref12,ref15, ref16,ref17,ref18,ref19,ref26,ref27, ref28, ref29, ref30, ref31, ref32, ref33,ref34, ref35, ref36, ref37,ref39, ref40, ref41, ref42,ref44, ref45,ref46,ref47,ref48,ref49,ref50,ref51}  \\ \hline 
Sensitivity (recall) &    $20$ & \cite{ref12, ref13, ref14, ref15,ref17,ref19,ref26, ref27, ref28,ref32, ref33, ref34, ref35, ref36,ref40,ref41,ref45,ref46,ref48,ref49}  \\ \hline 
AUC &   $17$ & \cite{ref12,ref15, ref16, ref17,ref19,ref26,ref29,ref30,ref32, ref33, ref34,ref36,ref40,ref42,ref47,ref48,ref50} \\ \hline 
Specificity &   $15$ & \cite{ref12,ref14,ref15,ref17,ref26, ref27, ref28,ref33, ref34, ref35, ref36,ref41,ref45,ref46,ref48} \\  \hline
Precision & $7$ & \cite{ref13,ref19,ref28,ref32,ref40,ref49 }  \\ \hline 
Positive predictive value (PPV) and Negative predictive value (NPV)   &  $3$ & \cite{ref15,ref28,ref35} \\ \hline 
Matthews correlation coefficient (MCC) &    $2$ & \cite{ref36, ref28} \\ \hline 
C-index &  $1$ & \cite{ref38} \\ \hline 
Root-Mean Squared Error (RMSE) &   $1$ & \cite{ref43} \\ \hline 
\multicolumn{3}{p{16cm}}{The numbers in the second column do not sum up to 34 as many studies used more than a single metric.} \\ \hline 
\end{tabular}
\caption{\label{tab:5}The distribution of evaluation metrics in the included studies.}
\end {table}

\section*{Discussion}
This section summarizes our findings and provides future directions for research on the multimodal fusion of medical imaging and EHR.

\subsection*{Principal findings}
We found that multimodal models that combined EHR and medical imaging data generally outperformed single modality models for the same task in disease diagnosis or prediction. Since our review shows that the fusion of medical imaging and clinical context data can improve the performance of AI models, we recommend attempting fusion approaches when multimodal data is obtainable. Moreover, through this review, we observed certain trends in the field of multimodality fusion in the medical area, which can be categorized as:
\begin{itemize}
    \item \textbf{\textit{Resources}}: We observed that multimodal data resources of medical imaging and EHR are limited owing to privacy considerations. The most prominent dataset was the ADNI, containing MRI and PET images collected from about 1700 individuals in addition to clinical and genetic information. Considering ADNI's contributions in advancing the research, similar multimodal datasets should be developed for other medical data sources too. 
    \item \textbf{\textit{Fusion implementation}}: Early fusion was the most commonly used technique in most applications for multimodal learning. Before fusing 1D EHRs data with image data in 2D or 3D, images data was converted to a 1D vector by extracting high-level representations using manual or software-generated features \cite{ref12,ref13,ref15,ref27,ref28,ref29,ref30,ref31,ref36,ref37,ref38,ref39,ref40,ref34,ref45,ref46,ref47,ref48}, or CNN-extracted features \cite{ref16,ref8,ref49}. The learned imaging features from CNN often  resulted in better task-specific performance than manually or software-derived features \cite{ref58}. Based on this reviewed studies, early fusion models performed better than conventional single-modality models on the same task. Researchers can use the early fusion method as a first attempt to learn multimodal representations since it can learn to exploit the interactions and correlations between features of each modality. Furthermore, it only requires one model to be trained, making the pipeline for training easier than that of joint and late fusion. However, if imaging features are extracted with CNN, early fusion requires multiple models to be trained.
    
    Joint fusion was the second most commonly used fusion approach. From a modality perspective, CNNs appeared to be the best option for image feature extraction. Tabular data were mainly processed using dense layers when fed into a model, while text data were mostly processed using LSTM layers followed by the attention layer. Most of the current research directly concatenated the feature vectors of the different modalities to combine multimodal data. Using NNs to implement joint fusion can be a limitation when dealing with small datasets, which means that joint fusion is preferred with large datasets. For small datasets, it is preferable to use early or late fusion methods as they can be implemented using classical ML techniques. Nevertheless, we expect and agree with \cite{ref21} that joint fusion models can provide better results than other fusion strategies because they update their feature representations iteratively by propagating the loss to all the feature extraction models, aiming to learn correlations across modalities. 
    
    Based on the performance reported in the included studies, it is preferred to try the early and joint fusion when the relation between the two data modalities is complementary. In this review, AD diagnosis is an example in which imaging and EHRs data are dependent as relevant and accurate knowledge of the patient's current symptomatology, personal information and imaging reports can help doctors interpret imaging results in a suitable clinical context, resulting in a more precise diagnosis. Therefore, all AD diagnosis studies in this review implemented either early fusion \cite{ref13,ref14,ref15} or joint fusion \cite{ref44} for multimodal learning.
    
    On the other hand, it is preferred to try late fusion when input modalities do not complement each other. For example, the brain MRI pixel data and the quantitative result of an MMSE (e.g., Qiu et al. \cite{ref32}) for diagnosing MCI are independent, making them appropriate candidates for inclusion in the late fusion strategy. Also, late fusion does not impose the requirement of a huge amount of training data, so it could be used when the modalities data sizes are small. Moreover, late fusion strategy could be attempted when the concatenation of feature vectors from multiple modalities results in high-dimensional vectors that are difficult for ML algorithms to learn without overfitting unless many input samples are available. In late fusion, multiple models are employed, each specialized in a single modality, thereby limiting the size of the input feature vector for each model. Furthermore, late fusion could be used when data is incomplete or missing, i.e., some patients have only imaging data but no clinical data or vice versa. This is because late fusion uses independent models for different modalities, and aggregation methods like averaging and majority voting can be used even when predictions from a modality are not present. Moreover, predictions could be disproportionately influenced by the most feature-rich input modality when the number of features is very different between the input data modalities \cite{ref59}; in this scenario, late fusion is preferable because it allows training each model using each modality separately.
    
    \item \textbf{\textit{Applications}}: In this review, we found that AD diagnosis and prediction \cite{ref12,ref13,ref14,ref15,ref39,ref43,ref44} were the most common applications addressed in a multimodal setting among studies. Using ML fusion techniques consistently demonstrated improved AD diagnosis, while clinicians experience difficulty with accurate and reliable diagnosis even when multimodal data is available \cite{ref21}. This emphasizes the utility and significance of multimodal fusion approaches in clinical applications.
    
    \item \textbf{\textit{Prospects}}: In this review, we noted that multimodal medical data fusion is growing due to its potential in achieving state-of-the-art performance for healthcare applications. Nonetheless, this growth is hampered by the absence of adequate data for benchmarking methods. This is not surprising, given the privacy concerns surrounding revealing healthcare data. Moreover, we observed a lack of complexity in the used non-imaging data, particularly in the context of heavily feature-rich data included in the EHR. For example, the majority of studies focused mostly on basic demographic data like gender and age \cite{ref12,ref15,ref39,ref46}, a limited number of studies also included medical histories such as smoking status and hypertension \cite{ref18,ref50} or specific clinical characteristics that are known to be associated with a certain disease, such as an MMSE for diagnosing AD. In addition to selecting the disease-associated features, future research may benefit from using vast amounts of feature-rich data, as demonstrated in domains outside of medicine, such as autonomous driving \cite{ref60}.
\end{itemize}

\subsection*{Future directions}
Although we focus on EHR and medical imaging as multimodal data, other modalities such as multi-omics and environmental data could also be integrated using the aforementioned fusion approaches. As the causes of many diseases are complex, many factors, including inherited genetics, lifestyle, and living environments, contribute to the development of diseases. Therefore, combining multisource data, e.g. EHR, imaging, and multi-omics data, may lead to a holistic view that can improve patient outcomes through personalized medicine. 

Although we focus on EHR and medical imaging as multimodal data, other modalities such as multi-omics and environmental data could also be integrated using the aforementioned fusion approaches. As the causes of many diseases are complex, many factors, including inherited genetics, lifestyle, and living environments, contribute to the development of diseases. Therefore, combining multisource data, e.g. EHR, imaging, and multi-omics data, may lead to a holistic view that can improve patient outcomes through personalized medicine. 

Moreover, the unavailability of multimodal public data is a limitation that hinders the development of corresponding research. Many factors (e.g., gender, ethnicity, environmental factors) could influence the research directions or even clinical decision, relying on a few publicly available datasets might not be enough for making conclusive clinical claims to the global population \cite{ref20}. Consequently, it is imperative to encourage the sharing of flexible data among institutions and hospitals in order to facilitate the exploration of a wider range of population data for clinical research.  In ML, federated learning (FL) \cite{ref61,ref62} provides the ability to collect data safely and securely from multiple centers. It may be used to collect multimodal data from various centers to train a large-scale model without collecting data directly. 

\subsection*{Limitations}
Our search was limited to studies published within the previous seven years (2015-2022). We only considered studies published in English, which may have led to leaving out some studies published in other languages. We solely included studies fusing EHR with medical imaging. We did not include studies that used other data modalities such as multi-omics data, as they are out of the scope of this work. Because positive results are typically reported disproportionately, publication bias might be another limitation of this review. This bias may result in an overestimation of the benefits associated with multimodal data analysis. The studies included in this review employed various input modalities, investigated various clinical tasks for different diseases, and reported different performance metrics; hence a direct comparison of the results presented in the studies is not always applicable. Furthermore, not all articles provided confidence bounds, making it difficult to compare their results statistically.

\section*{Conclusion}
Multimodal ML is an area of research that is gaining attention within the medical field. This review surveyed multimodal medical ML literature that combines EHR with medical imaging data. It discussed fusion strategies, the clinical tasks and ML models that implemented data fusion, the type of diseases, and the publicly accessible multimodal data for medical imaging and EHRs. Furthermore, it highlighted some directions to pave the way for future research. Our finding suggests that there is a growing interest in multimodal medical data. Still, most studies combine the modalities with relatively simple strategies, which despite being shown to be effective, might not fully exploit the rich information embedded in these modalities.  As this is a fast-growing field and new AI models with multimodal data are constantly being developed, there might exist studies that fall outside our definition of fusion strategies or use a combination of these strategies. We believe that the development of this field will give rise to more comprehensive multimodal medical data analysis and will be of great support to the clinical decision-making process.

\section*{Data availability}
The data generated during this scoping review is provided as supplementary materials.

\bibliography{main}

\section*{Author contributions statement}
F.M., H.A., Z.S. contributed to conceptualization. F.M. and H.A. administered the project. F.M. curated the data, performed data synthesis, and contributed to writing—original draft. H.A and N.E performed writing—review and editing. Z.S. and H.A. supervised the study. All authors read and approved the final manuscript.

\section*{Additional information} 
\textbf{Supplementary Methods}  \\
\textbf{Competing interests} \\
The authors declare that they have no competing interests. 


\end{document}